%% file: main.tex
\documentclass{article}
\usepackage{graphicx} %

\usepackage[a4paper, total={6in, 8in}]{geometry}

\usepackage{url}
\usepackage{hyperref}

\usepackage{amsmath}

\usepackage{xcolor}

\usepackage{cleveref}

\usepackage{listings}

\usepackage{csquotes}

\usepackage{nicefrac}

\usepackage{verbatim}

\newcommand{\abs}[1]{\left\lvert#1\right\rvert}

\usepackage{placeins}

\usepackage[style=authoryear]{biblatex}
\DeclareCiteCommand{\citeyear}
    {}
    {\bibhyperref{\printdate}}
    {\multicitedelim}
    {}

\makeatletter
\newcommand{\citet}[2][]{\citeauthor{#2} (\citeyear{#2}\ifthenelse{\equal{#1}{}}{}{, #1})}
\makeatother

\title{A dataset of rated conceptual arguments}

\author{Emery Cooper* \and Caspar Oesterheld* \and Linh Chi Nguyen \and Alexander Kastner \and Ethan Perez}
\date{September 19, 2025}

\begin{document}

\maketitle

\section{Introduction}

In recent years, the capabilities of large language models (LLMs) have progressed impressively across a wide range of domains.
Over the past year or so, so-called reasoning models %
have made rapid progress on tasks with verifiable feedback, such as coding and math \parencites{jaech2024openai,guo2025deepseek,yang2025qwen3}.

In light of concerns about the risks of AI development (such as misalignment \parencites[e.g.,][]{bostrom2014superintelligence}, mass unemployment and concentration of power%
, and gradual disempowerment \parencite{kulveit2025gradual}), a variety of authors have proposed agendas of differential acceleration: making models better at specific skills that are likely to have positive impact on the world and perhaps ones that aren't closely tied to general capabilities. Proposals include using AI specifically for AI safety \parencites[e.g.,][]{carlsmithAIforAIsafety}, using AI for science \parencites[e.g.,][]{bengioscience}, and improving AI's ability to achieve cooperation and avoid conflict in multi-agent situations \parencites{dafoe2020open,clifton2020cooperation,critch2020ai,conitzer2023foundations,hammond2025multi}. %

In this project, we want to specifically work toward making AI better at (helping humans with) reasoning about what we call \textit{conceptual} questions. By this, we roughly mean questions with two properties:
\begin{itemize}
    \item We have no (realistically accessible) ground truth answer to the question; and no widely accepted methodology for resolving the question.
    \item We can make progress by considering and debating arguments.
\end{itemize}
Most philosophical issues (e.g., ethics, the nature of free will and consciousness) are prime examples of conceptual questions. Mathematical questions (e.g., \enquote{is there a polynomial-time prime factorization algorithm}, \enquote{is the Riemann conjecture true?}) and many empirical questions (e.g., \enquote{can chemical compound X cause cancer?}) are paradigmatic examples of non-conceptual questions. Many important areas mix conceptual and non-conceptual issues. For instance, social aggregation of preferences might involve conceptual questions such as: ``Does formalization X faithfully represent intuitive concept Y?'', ``In light of the multiplicity of equilibria, what is a `good'/`fair' equilibrium in this situation?'', ``What properties should a voting rule have?'', etc. At the same time, it also involves various non-conceptual questions, such as whether a given set of voting-theoretic axioms is mutually inconsistent.

\subsection{A high-level approach: multi-dimensional evaluation of contextualized arguments}
\label{sec:high-level-approach}

The main obstacle to improving LLMs' conceptual reasoning is that we don't have access to ground truth on conceptual questions. E.g., we don't know whether utilitarianism is the best normative ethical theory, whether humans have free will, whether GPT-5 is conscious, etc.

Our approach to circumventing this problem is based on the following theses:
\begin{enumerate}
    \item It is possible to evaluate \emph{arguments} about conceptual topics.
    \begin{enumerate}
        \item On philosophical issues, it is easier (less contentious and subjective) to evaluate arguments than it is to evaluate bottom-line conclusions.
        \item It's easier to evaluate \textit{contextualized} arguments. That is, it's easier to evaluate an argument if it is placed in context of existing arguments, considerations, claims, and theories.\label{thesis:contextualized}
        \item Sometimes it's easier to evaluate arguments on specific dimensions -- e.g., how central they are to a particular issue -- than holistically.
    \end{enumerate}
    \item Evaluating arguments is useful to make progress on bottom-line conclusions on conceptual topics.
\end{enumerate}
    
To illustrate the first theses, consider the following example. It's hard to say whether utilitarianism is a good (or the best) moral theory. But we can, to some extent, evaluate arguments for or against utilitarianism, especially if we are given a context of existing arguments, considerations, claims, and theories. For instance, let's say Alice proposes utilitarianism as a moral theory. %
Now consider the following possible responses by Bob:
\begin{enumerate}
    \item \label{item-contra-utilitarianism:trolley} Bob could argue that consequentialism -- the view that how good an action is is determined solely by its consequences, such as the social good in Alice's proposal -- is not very compelling. Bob could give examples like the famous Trolley problem \parencites{foot1967problem,thomson1976killing}, in which our intuitive moral attitudes toward an action seem to depend not just on the consequences of the action. 
    \item \label{item-contra-utilitarianism:infinite-ethics} Bob could argue that utilitarianism struggles with cases of infinite utilities \parencites[for overviews of these issues, see][]{bostrom2011infinite,arntzenius2014utilitarianism, Askell2018}. %
\end{enumerate}
Regardless of what we ultimately make of utilitarianism, it seems like we can say the following about these arguments: Both are sensible counterarguments, assuming they are unaddressed in Alice's opening statement. Meanwhile, if Alice, say, explicitly advocates utilitarianism in a finite setting, then Objection \ref{item-contra-utilitarianism:infinite-ethics} becomes (nearly) irrelevant; similarly, if Alice explicitly addresses the counterintuitive nature of utilitarianism's recommendation, then minimal versions of Objection \ref{item-contra-utilitarianism:trolley} become weak. We might further say that Objection \ref{item-contra-utilitarianism:trolley} is a more \textit{central} critique of utilitarianism than Objection \ref{item-contra-utilitarianism:infinite-ethics}. Moreover, it's easy to give versions of Objection \ref{item-contra-utilitarianism:infinite-ethics} that are very solid (hard to argue with), whereas Objection \ref{item-contra-utilitarianism:trolley} will always rely on appeals to intuition. %

It's hard to give a theoretical justification for our first theses, but we can cite empirical evidence. For example, few high-level philosophical debates have ever been settled. But many philosophical \emph{arguments} are widely acknowledged to be important by both sides of the respective debates. Consequently philosophy textbooks generally focus on teaching arguments. Some aspects of our own project can be seen as assessing (and supporting) the first theses (see \Cref{appendix:rating-validation,sec:rating-issues}). Meanwhile, our second thesis is supported by the observation that people's typically construct and consider arguments to reach bottom-line conclusions on philosophical topics.

\subsection{Introducing a dataset on conceptual reasoning}

We created a dataset consisting of position texts (\emph{positions}), argumentative \emph{critiques} of the position texts, and human-expert ratings of the critiques along the following dimensions: centrality (how central the issue attacked by the critique are to the position text), strength (how successfully the critique attacks these issues), correctness (to what extent the critique makes correct, rather than false claims),  clarity (how unambiguous the critique is after careful reading), and \enquote{dead weight} (how much of the critique is unrelated to criticizing the position text). Separately from the above indicators of the quality of the critique, we also assess \enquote{single issue} (to what extent the critique focuses on a single issue in the position text). Finally, we give each critique a holistic overall rating. We give some examples of data points in \Cref{table:example:aunt-dahlia,table:example:voting}. We have \input{current_data_stats/no_of_dialogues_with_ratings} rated critiques -- see \Cref{table:dataset-stats} for more details.
The dataset contains arguments on a variety of topics including AI safety, decision theory, normative ethics, philosophy of mind, politics, and various miscellaneous topics.
We describe the dataset in more detail in \Cref{sec:dataset-details}.

In \Cref{sec:suggested-scoring-functions}, we describe two scoring functions we use for the dataset, one ranking-based loss based on pairs of critiques of the same position and one point-wise loss.

In \Cref{sec:basic-experiments}, we assess performance of a few modern models with chain of thought. We find that scores on our datasets align well with the models' general capabilities. However, we also find that ``thinking''/``reasoning'' doesn't typically improve performance.

Finally, we discuss related work in \Cref{sec:related-work}.

\section{Details on our dataset}
\label{sec:dataset-details}

\begin{table}[]
    \centering
    \begin{tabular}{lc}
        \hline
        \hline
            Number of (rated) dialogues/critiques & \input{current_data_stats/no_of_dialogues_with_ratings}\\
            \hline
            Number of ratings (ignoring revisions) by ... & \\
            \quad ... Emery Cooper & \input{current_data_stats/no_of_dialogues_rated_by_Emery_Cooper}\\
            \quad ... Caspar Oesterheld & \input{current_data_stats/no_of_dialogues_rated_by_Caspar_Oesterheld}\\
            \quad ... Alexander Kastner & \input{current_data_stats/no_of_dialogues_rated_by_Alex}\\
            \quad ... Linh Chi Nguyen & \input{current_data_stats/no_of_dialogues_rated_by_Chi}\\
            \quad ... Lukas Gloor & \input{current_data_stats/no_of_dialogues_rated_by_Lukas_Gloor}\\
            \quad ... Lukas Finnveden & \input{current_data_stats/no_of_dialogues_rated_by_Lukas_Finnveden}\\
            Total & \input{current_data_stats/total_no_of_ratings}\\
            \hline
            Number of (rated) model-written critiques & \input{current_data_stats/n_model_written_critiques}\\
        \hline
            Number of positions (with at least one rated critique) & \input{current_data_stats/no_of_arguments_with_geq_1_rated_critiques}\\
            Number of positions with $\geq 2$ (rated) critiques & \input{current_data_stats/no_of_arguments_with_geq_2_rated_critiques}\\
        \hline
        \hline
    \end{tabular}
    \caption{An overview of the size of our dataset}
    \label{table:dataset-stats}
\end{table}

\begin{table}[]
    \centering
    \input{examples/bad_100_86b_1_3}
    \caption{An example from our dataset}
    \label{table:example:aunt-dahlia}
\end{table}

\begin{table}[]
    \centering
    \input{examples/le_chat_voting_1_4}
    \caption{An example from our dataset}
    \label{table:example:voting}
\end{table}

\paragraph{Generating positions and critiques} We obtained a diverse dataset of positions and critiques by using a wide range of sources: some are hand-written (27 positions, a few with LLM assistance), some are adapted from books (168 positions), %
some are extracted from volunteers' philosophy coursework (41 positions, some with LLM assistance), others are from
magazines, blogs or online forums (108 positions), some are primarily model-written (68 positions), and a few are adapted from other datasets (28 positions).

Additionally, we generated some critiques by using LLMs with an active learning-like approach. %
Specifically, we generated a large number of critiques using LLMs, then let some of our best LLM-based judges rate all the critiques, and then added those critiques to the dataset on which the best judges disagree. This way, we automatically generate critiques that are difficult for the best judges (and best models). We also added some LLM-generated critiques that were highly rated by the best judges (including both correctly highly rated and erroneously highly rated critiques). In both cases, the critiques were ultimately hand-selected for diversity, suitability and interestingness.

\paragraph{Rating the critiques} We rated the critiques by hand. The full rubric describing the meaning of the different fields is given in \Cref{appendix:the-full-rubric}. Source and tags aren't shown when rating a critique.\footnote{Some early ratings were created with visible sources and tags. However, source is an obvious confounder and was thus made invisible relatively early on. Many tags -- e.g., tags for topics, such as ``decision theory'' -- aren't confounders. But partway through data generation we started using tags that that would confound ratings, and thus had to remove them from view.} By default, critiques of the same position are shown in sequence. However, for various reasons critiques of the same position may still be rated at different times.\footnote{For one critiques of the same position may be created at different times. So in particular, one critique may be created after other critiques are already rated. Second, rating may be interrupted while going through one position's critiques.} Time per rating varies with the length of the position and critique. For short position/critique pairs, rating takes about 5--15 minutes on average. Almost all of our data points are at least rated by Emery Cooper. Some are also rated by Caspar Oesterheld, Alexander Kastner, Linh Chi Nguyen, Lukas Finnveden and Lukas Gloor.

To increase the quality of our dataset, we revisit ratings according to the following methods. First, when multiple experts rate the same position very differently (e.g., high difference in overall scores), they discuss their disagreement. If one of the experts makes a point that has been overlooked by the other (e.g., a possible interpretation of the position), the other expert may add a new rating, indicating the cause of the revision in the comments. Note that the original rating is always preserved. We conducted a test to determine how well ratings prior to discussion predict ratings after discussion, which we discuss in \Cref{appendix:rating-validation}.

Second, our primary rater Emery Cooper checked her biggest disagreements against the strongest LLM-based raters. %

\paragraph{Limitations}
It is important to keep in mind that there are distributional differences between good and bad critiques in our dataset. For instance, model-generated critiques are disproportionately weak. Long critiques that we wrote ourselves (and are thus written in a distinct writing style) are almost always strong. Consequently, we suspect it's not that difficult to achieve strong performance on the dataset in meaningless ways by picking up on these distributional differences. Relatedly, it should be noted that while we rated the critiques procedurally blind to the source, it is often obvious whether a critique is, say, written by an LLM. 

Additionally, there are currently relatively few positions with a good spread of good and bad critiques. There are currently a lot of positions with only one critique, or with only bad critiques. This poses a challenge for scoring models based on loss functions that compare models' rankings of critiques within a given position to the human rankings of those same critiques.

Finally, it is important to note that there is still some subjectivity involved in the human ratings. We discuss this in more detail in \Cref{sec:rating-issues} and \Cref{appendix:rating-validation}.

\section{Scoring functions}
\label{sec:suggested-scoring-functions}

We score the performance of models by comparing their ratings to human ratings. The choice of performance metric for this dataset requires some care.  Using the wrong metric could render the dataset much less useful. The main considerations to be aware of are as follows: (a) strength and centrality should not be used in isolation, only in the form of the product strength*centrality, due to ambiguity in allocation between strength and centrality (see \Cref{sec:rating-issues}), (b) for some metrics, calibration significantly affects the score, so it may be necessary to recalibrate the models, (c) metrics that are based on comparisons of critiques within a position still give somewhat noisy results, due to the relatively small number of positions with a good spread of good and bad critiques.

We describe these issues in \Cref{sec:obstacles}. We are not yet sure ourselves what the best metric is. We briefly describe the metrics we use for the results here below, and formally define them and describe them in more detail in \Cref{sec:loss-fns}.

Finally, note that to properly understand the ratings, it is important to read the rubric, since the meaning of different scores is specified precisely there, and may differ from the intuitive meaning.

Our first metric is the \emph{weighted pairwise ranking error rate}. Essentially, on each position, we take the proportion of comparisons of critiques that the model gets wrong, weighted by the difference in human ratings for those critiques. That is, we look at each possible comparison of critiques, and give a loss of zero if the model's overall scores have the same order as the human's, and otherwise give a loss proportional to the difference in the \emph{human} overall scores for that comparison. We then divide by the total number of possible comparisons on this position. Roughly, this is the proportion of comparisons that the model gets wrong, weighted by how bad each error is. We then average across all positions in the dataset. We describe this formally in \Cref{sec:loss-fns}.

Our other metric is the \emph{custom weighted loss}. Essentially, this is a weighted average of absolute errors on each rubric item (strength, centrality, correctness, etc.), with two modifications: First, if the human rates the clarity as very low, only clarity and overall are used. This is because the other ratings are likely to be unreliable if the human cannot understand what the critique is saying. Second, we combine strength and centrality into a single metric, their product, rather than considering these scores individually. We describe this in more detail in \Cref{sec:loss-fns}.

\section{Some basic experiments}
\label{sec:basic-experiments}

\paragraph{Example model failure}
 Consider \Cref{table:example:aunt-dahlia}, Critique 1. This table shows Opus 4's response on this example (one of its largest mistakes on the dataset). This is a fairly typical instance of a model failure: the model comes to a wrong conclusion without actually thinking very much. We found that with e.g. few-shot prompting, models can be encouraged to reason much more thoroughly, avoiding some (but not all) such failures.

\begin{table}[]
    \begin{tabular}{{p{\linewidth}}}
\hline
    Let me analyze this critique step by step.

**Understanding the Position:**
The position claims to demonstrate that one can infer \enquote{ought} from \enquote{is}, thereby bridging the is-ought gap. It presents a logically valid argument where:
- Premise 1 describes what someone believes (including a moral claim)
- Premise 2 describes the truth-value of those beliefs
- The conclusion is a moral proposition

The position's central claim is that this valid argument bridges the is-ought gap because both premises are descriptive (\enquote{is} statements) while the conclusion is normative (\enquote{ought} statement).

**Understanding the Critique:**
The critique makes two points:
1. Acknowledges the argument is logically valid
2. Claims it doesn't bridge the is-ought gap because Premise 1 already contains an \enquote{ought}

**Centrality (0-1):**
The critique directly attacks the core claim - that this argument bridges the is-ought gap. If the critique is correct that Premise 1 already contains an \enquote{ought,} then the position completely fails in its purpose. The entire point of the position is to show bridging from \enquote{is} to \enquote{ought,} so attacking whether this bridging actually occurs is maximally central.
**Centrality: 1.0**

**Strength (0-1):**
The critique makes a strong point. Premise 1 states \enquote{\enquote{Bertie morally ought to marry Madeline} is one of Aunt Dahlia's beliefs.} This premise describes a belief, but the content of that belief is itself an ought-statement. The is-ought problem is about deriving normative conclusions from purely descriptive premises. While Premise 1 is grammatically descriptive (it describes what someone believes), it smuggles in normative content through the belief's content. This is a well-known fallacy in attempts to bridge the is-ought gap - you can't truly derive an ought from pure is-statements if your is-statements already contain oughts embedded within them. The critique effectively refutes the position's claim.
**Strength: 0.9**

**Correctness (0-1):**
The critique makes two claims:
1. The argument is logically valid - this is correct
2. Premise 1 already contains an \enquote{ought} - this is also correct (the belief described contains an ought-statement)

Both statements are true and accurately identify the problem with the position.
**Correctness: 1.0**

**Clarity (0-1):**
The critique is extremely clear and concise. It acknowledges what the position gets right (logical validity) and precisely identifies what it gets wrong (Premise 1 contains an ought). There's no ambiguity about what the critique is claiming.
**Clarity: 1.0**

**Dead Weight (0-1):**
The critique contains no extraneous material. Every part contributes to its argument. The acknowledgment of logical validity isn't dead weight because it shows the critique understands the position and focuses its attack precisely.
**Dead Weight: 0.0**

**Single Issue (0-1):**
The critique focuses on exactly one issue: whether the argument truly bridges from \enquote{is} to \enquote{ought} given that Premise 1 contains an ought-statement. It doesn't raise multiple independent objections.
**Single Issue: 1.0**

**Overall (0-1):**
This is an excellent critique. It identifies a fundamental flaw in the position with precision and clarity. The strength*centrality product is 0.9, indicating it nearly completely refutes the position. The critique is insightful (recognizing the hidden ought in Premise 1), perfectly clear, contains no errors, and no extraneous material. It's exactly the kind of focused, devastating critique that philosophical arguments sometimes face when they try to solve classical problems through linguistic sleight-of-hand.
**Overall: 0.9**\\
\hline
    \end{tabular}
    \caption{An example of an LLM's rating of \Cref{table:example:aunt-dahlia}, critique 1}
    \label{tab:placeholder}
\end{table}

\paragraph{Ranking the performance of some models}
Are better models also better at judging philosophical arguments in particular? One would certainly expect that more powerful base models also have greater philosophical abilities, but it might take some effort to elicit them. After all, the pretraining data (presumably) contains both high-quality and low-quality conceptual reasoning.
So what happens with a simple baseline prompt (i.e., one that doesn't contain examples, etc.)?
Without knowing how companies chat/instruct-tune/RLHF their models, it is not clear what behavior we should expect. Better models can have have worse judgment if, for example, they're more aggressively trained to appease human users, as demonstrated by the phenomenon of sycophancy. %
Similarly, safety training could make models worse at questioning various widely accepted ideas, an essential philosophical skill.

In \Cref{table:model-ranking} we give a ranking of models in terms of the weighted pairwise ranking error (as described in \Cref{sec:hinge-comparison-loss}) against Emery Cooper's ratings. We use a simple baseline prompt, see \Cref{appendix:baseline-pairwise-comparison-prompt}.

The table shows that -- contrary to the above worries -- our ranking of models aligns with a general sense of how good the different models are. Generally, within company, model rankings align with expectations: for a given tier of models, newer models are typically better (GPT 5 is better than GPT 4.1, Claude Opus 4 is better than Claude Opus 3, etc.); and within model generations, the higher-tiered models are better (Claude Opus is better than Sonnet, Gemini Pro is better than Flash, etc.). %
Our ranking also aligns with other rankings.\footnote{For example, consider SimpleBench (\citeyear{simplebench}) with scores listed at \url{https://simple-bench.com/} or the leaderboard at \url{https://artificialanalysis.ai/leaderboards/models} (Methodology explained at \url{https://artificialanalysis.ai/methodology/intelligence-benchmarking}), which aggregates a number of different benchmarks; or even the popular but arguably less meaningful LMArena leaderboard at \url{https://lmarena.ai/leaderboard/text}.
}

Meanwhile, we find that so-called ``reasoning'' or ``thinking''\parencites{jaech2024openai,guo2025deepseek,yang2025qwen3}, doesn't help models much in judging critiques. In \Cref{table:thinking-versus-non-thinking}, we specifically list the scores of pairs of the same model with versus without thinking (varying either an API parameter or the model itself). ``Thinking'' sometimes helps and sometimes hurts performance. The difference is generally small (e.g., relative to 95\% confidence intervals).

We believe that longer reasoning can help with getting higher scores, so perhaps it is surprising that ``thinking'' doesn't help.  Our best guess is that ``thinking'' is primarily trained in specific domains (math, coding) with easily verifiable answers and that training on those domains doesn't generalize to the more conceptual issues discussed in this paper.

\begin{table}[]
    \centering
    {
    \begin{tabular}{lrr}
\hline
Judge                                              &   Avg Loss &   95\% CI \\
\hline
 gpt-5-2025-08-07                                &      0.080 &    0.016 \\ 
 o3-pro-2025-06-10                               &      0.083 &    0.018 \\ 
 claude-opus-4-1-20250805                        &      0.090 &    0.017 \\ 
 o3-2025-04-16                                   &      0.094 &    0.020 \\ 
 claude-sonnet-4-20250514                        &      0.098 &    0.016 \\ 
 claude-opus-4-20250514                          &      0.099 &    0.017 \\ 
 gpt-5-mini-2025-08-07                           &      0.103 &    0.021 \\ 
 claude-3-5-sonnet-20241022                      &      0.103 &    0.017 \\ 
 o4-mini-2025-04-16                              &      0.104 &    0.020 \\ 
 claude-3-7-sonnet-20250219                      &      0.111 &    0.019 \\ 
 gemini-2.5-pro                                  &      0.115 &    0.017 \\ 
 gemini-2.5-flash                                &      0.118 &    0.018 \\
 o1-2024-12-17                                   &      0.122 &    0.021 \\ 
 gpt-4.1-2025-04-14                              &      0.127 &    0.019 \\ 
 gpt-4.1-mini-2025-04-14                         &      0.128 &    0.022 \\ 
 gpt-oss-120b-Q5-K-S                             &      0.131 &    0.024 \\ 
 Qwen3-30B-A3B-Thinking-2507-Q6-K                &      0.138 &    0.020 \\ 
 Qwen3-30B-A3B-Instruct-2507-Q6-K                &      0.139 &    0.022 \\ 
 magistral-medium-2506                           &      0.141 &    0.022 \\ 
 Qwen3-4B-Instruct-2507-Q8-0                     &      0.142 &    0.017 \\ 
 gemini-2.0-flash-001                            &      0.143 &    0.022 \\ 
 mistral-medium-2505                             &      0.145 &    0.022 \\ 
 gpt-4o-2024-08-06                               &      0.146 &    0.024 \\ 
 gemini-2.5-flash-lite                           &      0.146 &    0.022 \\ 
 claude-3-opus-20240229                          &      0.148 &    0.024 \\ 
 gemini-1.5-pro-002                              &      0.148 &    0.023 \\ 
 Qwen3-4B-Thinking-2507-Q8-0                     &      0.149 &    0.021 \\ 
 magistral-small-2506                            &      0.150 &    0.020 \\ 
 gpt-5-nano-2025-08-07                           &      0.152 &    0.021 \\ 
 gpt-4-0125-preview                              &      0.152 &    0.024 \\ 
 Qwen3-30B-A3B-Q6-K                              &      0.152 &    0.022 \\ 
 gpt-oss-20b-Q5-K-S                              &      0.154 &    0.026 \\ 
 claude-3-5-haiku-20241022                       &      0.155 &    0.022 \\ 
 gpt-3.5-turbo-1106                              &      0.158 &    0.024 \\ 
 phi-4-Q5-K-M                                    &      0.162 &    0.022 \\ 
 gemma-3-27b-it.Q8-0                             &      0.163 &    0.026 \\ 
 Qwen3-8B-Q6-K                                   &      0.164 &    0.023 \\ 
 mistral-small-2503                              &      0.165 &    0.024 \\ 
 claude-3-haiku-20240307                         &      0.168 &    0.021 \\ 
 gemini-1.5-flash-002                            &      0.171 &    0.024 \\ 
 random                                          &      0.173 &    0.016 \\ 
 gpt-4.1-nano-2025-04-14                         &      0.173 &    0.023 \\ 
 gemini-2.0-flash-lite-001                       &      0.174 &    0.025 \\
\hline
    \end{tabular}
    }
    \caption{A ranking of models in terms of the weighted pairwise ranking error rate. Loss is aggregated across 255 arguments and 856 critique pairs.
    }
    \label{table:model-ranking}
\end{table}

\begin{table}
\centering
{
    \begin{tabular}{lrr}
\hline
Judge                                              &   Avg Loss &   95\% CI \\
\hline
 claude-opus-4-1-20250805-thinking10k            &      0.086 &    0.018 \\ 
 claude-opus-4-1-20250805                        &      0.090 &    0.017 \\ 
&&\\
 claude-opus-4-20250514-thinking10k              &      0.094 &    0.018 \\ 
 claude-opus-4-20250514                          &      0.099 &    0.017 \\ 
&&\\
 claude-sonnet-4-20250514                        &      0.098 &    0.016 \\ 
 claude-sonnet-4-20250514-thinking10k            &      0.112 &    0.021 \\ 
&&\\
 claude-3-7-sonnet-20250219                      &      0.111 &    0.019 \\ 
 claude-3-7-sonnet-20250219-thinking10k          &      0.111 &    0.019 \\ 
&&\\
 gemini-2.5-flash-no-thinking                    &      0.112 &    0.020 \\ 
 gemini-2.5-flash                                &      0.118 &    0.018 \\
&&\\
 gemini-2.5-flash-lite-dynamic-thinking          &      0.135 &    0.021 \\ 
 gemini-2.5-flash-lite                           &      0.146 &    0.022 \\ 
&&\\
 Qwen3-30B-A3B-Thinking-2507-Q6-K                &      0.138 &    0.020  \\ 
 Qwen3-30B-A3B-Instruct-2507-Q6-K                &      0.139 &    0.022 \\ 
&&\\ 
 magistral-medium-2506                           &      0.141 &    0.022 \\ 
 mistral-medium-2505                             &      0.145 &    0.022 \\ 
&&\\
 Qwen3-4B-Instruct-2507-Q8-0                     &      0.142 &    0.017 \\ 
 Qwen3-4B-Thinking-2507-Q8-0                     &      0.149 &    0.021 \\ 
&&\\
 magistral-small-2506                            &      0.150 &    0.020  \\ 
 mistral-small-2503                              &      0.165 &    0.024 \\ 
\hline
    \end{tabular}
    }

    \caption{Comparing pairs of models with and without thinking on the weighted pairwise ranking error rate. Note that in gemini-2.5-flash-lite reasoning is turned off by default, while gemini-2.5-flash's default is ``dynamic reasoning''.}
    \label{table:thinking-versus-non-thinking}
\end{table}

\begin{table}[]
    \centering
    \begin{tabular}{lrr}
\hline
Judge                                                                &   Avg Loss &   95\% CI \\
\hline
 gpt-5-2025-08-07           &      0.211 &    0.009 \\
 claude-opus-4-1-20250805   &      0.222 &    0.010 \\
 o3-pro-2025-06-10          &      0.234 &    0.009 \\
 o3-2025-04-16              &      0.236 &    0.009 \\
 claude-opus-4-20250514     &      0.242 &    0.011 \\
 o4-mini-2025-04-16         &      0.276 &    0.014 \\
 claude-sonnet-4-20250514   &      0.279 &    0.012 \\
 claude-3-5-sonnet-20241022 &      0.290 &    0.013 \\
 gpt-5-mini-2025-08-07      &      0.291 &    0.013 \\
 claude-3-7-sonnet-20250219 &      0.301 &    0.013 \\
 gpt-5-nano-2025-08-07      &      0.312 &    0.010 \\
 gemini-2.0-flash-001       &      0.318 &    0.013 \\
 gemini-2.5-pro             &      0.319 &    0.016 \\
 o1-2024-12-17              &      0.321 &    0.015 \\
 gpt-4.1-2025-04-14         &      0.338 &    0.015 \\
 gpt-4.1-mini-2025-04-14    &      0.356 &    0.015 \\
 gemini-1.5-pro-002         &      0.356 &    0.016 \\
 claude-3-5-sonnet-20240620 &      0.357 &    0.015 \\
 gemini-2.0-flash-lite-001  &      0.358 &    0.015 \\
 claude-3-opus-20240229     &      0.360 &    0.015 \\
 mistral-small-2503         &      0.365 &    0.015 \\
 gpt-4o-2024-08-06          &      0.365 &    0.015 \\
 gemini-1.5-flash-002       &      0.374 &    0.015 \\
 gemini-2.5-flash           &      0.375 &    0.016 \\
 mistral-medium-2505        &      0.379 &    0.016 \\
 claude-3-5-haiku-20241022  &      0.382 &    0.015 \\
 gemini-2.5-flash-lite      &      0.409 &    0.017 \\
 claude-3-haiku-20240307    &      0.413 &    0.015 \\
 gpt-4.1-nano-2025-04-14    &      0.433 &    0.016 \\
\hline
    \end{tabular}
    \caption{A ranking of models in terms of the custom metric described in \Cref{sec:custom-loss}, averaged over 933 dialogues.}
    \label{tab:placeholder}
\end{table}

\section{Related work}
\label{sec:related-work}

\paragraph{Existing work on debate}
A large literature has developed around the idea of having AIs debate with each other in order to enable humans to make better decisions or, similarly, to allow humans to efficiently oversee LLM decision making \parencite{irving2018ai}.
The central objective of our project can be viewed as enabling AI for debate of conceptual issues. 

Like other streams in the safety for debate literature, our project is based on the assumption that being supplied with arguments is helpful for human decision making %
(see \Cref{sec:high-level-approach}). In case of debate on ground truth topics, this hypothesis can be studied both theoretically \parencites[e.g.,][]{irving2018ai,Barnes2020,brown2023scalable}
and experimentally \parencites[e.g.,][]{Barnes2020,khan2024debating,lang2025debate}.
Compared to other domains, it is less clear what alternative -- other than not using AI at all -- there are for philosophy. In domains with uncontroversial truth criteria, alignment alone (without oversight, human intervention) is in principle sufficient for powerful AI systems to reach correct judgment.

That said, a lot of assumptions, obstacles and ideas in the debate literature are relevant for the present direction as well. For instance, the problem of obfuscated arguments \parencite{Barnes2020obfuscated} is just as relevant (if not more relevant) in philosophy as it is in other areas. %
The ``clarity'' rating in our rubric is included in part to address this problem.

\paragraph{Argumentation datasets}

Most similarly, \citeauthor{tiwari2025debatebench}'s (\citeyear{tiwari2025debatebench}) DebateBench provide a dataset of transcribed competitive debates along with competitive debate judges' ratings of the individual speeches in the debate. We include 18 positions adapted from DebateBench in our dataset.

The VivesDebate \parencites{ruiz2021vivesdebate,ruiz2023vivesdebate} is similarly constructed by transcribing competitive debates and jury ratings. However, the debates underlying VivesDebate are originally in Catalan (and are thus machine-translated into English) and all concern the topic of whether gestational surrogacy should be legalized. We include 6 positions adapted from VivesDebate in our dataset.

An early dataset of argument quality/persuasiveness rankings was given by \citet{Habernal2016argument}.
Numerous similar argumentation datasets were also created as part of IBM's Project Debater around 2020 \parencite{slonim2021autonomous}\footnote{For an overview of all the datasets, see \url{https://research.ibm.com/haifa/dept/vst/debating_data.shtml}.}, including datasets of rated and ranked arguments \parencites{Gleize2019convinced,toledo-etal-2019-automatic,gretz2020large}. %
In all cases, each data point consists of a claim and an argument in favor or against the claim, as well as crowd-worker-generated (unidimensional) ratings of the claims; or a \textit{pairwise comparison} of arguments for the same claim. Unfortunately, arguments are very short (1-2 sentences) and thus relatively shallow, perhaps more optimized for AI capabilities of its time than the current dataset.

\citet{Jin2022logical} introduce a dataset for detecting a specific set of fallacies (such as circular reasoning) in texts \parencite[see also][]{lim2024evaluation}. Pointing out fallacies can be seen as a form of critique. The more challenging part of their dataset is focused specifically on (empirical issues related to) climate change.

A few prior works and datasets have focused on critiques in different kinds of domains or with different or no truth criteria. For instance, \citet{saunders2022self} study critiquing summaries. \citet{recchia2025findtheflaws}, \citet{mcaleese2024llm}, \citet{wang2025critique} %
focus on domains with uncontroversial ground truths like computer programming.
\citet{roush2024opendebateevidence}, \citet{Cheng2022iam}, \citet{roush2020debatesum} and other works have introduced datasets of arguments without including any ratings of the arguments/speeches.
\citet{blanchard2013toefl11} provide a (as far as we can tell not publicly available) dataset of expert-rated essays. At least some of the essays are argumentative, but the ratings primarily assess language ability.

Some questions in the LSAT (the US-American Law School Admission Test) are of the form: ``Which of the following is assumed by the above text?'' or ``Which of the following, if true, would refute the above argument?''  Prior work has used LSAT-based datasets for NLP \parencites{wang2022lsat,zhong2022analytical}. These questions are essentially about assessing the ``centrality'' dimension (in a binary way -- i.e., between 0 and 1) of a hypothetical critique. In fact, we have generated a few data points in our data set from LSAT questions. That said, the LSAT questions usually aren't in suitable domains.

Various prior works have studied the \textit{persuasiveness} of LLM-written versus human-written arguments to \textit{non}-experts \parencites{durmus2024persuasion,schoenegger2025large,reddit2025persuasiveness}.

\paragraph{Different kinds of datasets on similar topics} Other datasets have been provided on related topics (e.g., philosophy, ethics) but with quite different methodologies, e.g., focusing on factual questions (what does theory X recommend?), taking a descriptive approach to normative questions (does model agree more with approach A or approach B?); e.g., see \citet{perez2022discovering}, \citet{feng2023pretraining}, \citet{oesterheld2024dataset}, \citet{barkhordar2024unexpected}, \citet{nie2023moca}, \citet{santurkar2023whose}.

\section*{Acknowledgments}

We thank Lukas Gloor and Lukas Finnveden for helpful discussions and ratings of some of our critiques. %

\printbibliography

\appendix 

\section{Obstacles for scoring functions}
\label{sec:obstacles}

Here, we discuss why some measures might give unsatisfactory results if applied naively, and general challenges for measures of loss on this dataset.

\paragraph{Sensitivity of measures to priors and calibration}
Some loss functions, particularly the mean squared error (MSE), are very sensitive to a model's calibration or how extreme vs middling its ratings tend to be. For instance, if all human ratings were binary (0 or 1), a model with 80\% accuracy would perform much better if it rated things 0.17 and 0.83 instead of 0 and 1. In general, if models are somewhat inaccurate, models are rewarded for shrinking their ratings towards the middle of the range. A weak model may even do best to give almost everything 0.5. In fact, only a few frontier models have a lower MSE on the overall score than always giving the dataset's average overall score (of a bit over 0.3). It's not clear that we want this behavior or care about measuring it, however, particularly if we are mostly interested in how models order the critiques.

Moreover, measures like like mean squared error (MSE) and absolute error are sensitive to the model's priors about critique quality. Imagine a model that has little understanding of the arguments and critiques and thus gives (roughly) the same score to most ratings. We can see this score as the model's prior expected critique quality, i.e., the model's guess about how good the critique is without having assessed the critique. Then the model's loss on many metrics will be highly sensitive to what this prior expected critique quality is. But having an accurate prior about the critique quality (i.e., accurately expecting the average overall score to be around 0.3 rather than, say, 0.6) is not a ``skill'' (or feature) that we are particularly interested in. So some metrics in large part measure an irrelevant feature of the models.

To avoid these issues, it may be necessary for some purposes to recalibrate models to give similar score distributions to humans prior to ranking them.

\paragraph{Noise} Even if all human ratings are correct (in the sense of being based on a full understanding) and even if everyone (models and human raters) are calibrated to each other and have same prior expectations, the fuzziness of the exact ratings still introduces noise. This noise threatens to crowd out the genuine signal.

As an example, imagine we have a single position with many critiques. Let's imagine that almost all critiques are bad (overall score near 0) and only one critique is good (overall score near 1). Intuitively, it seems most important that a model identifies the good critique as good and the bad critiques as bad. But for most natural loss functions, some of the signal will come the exact ratings of a lot of the weak critiques. In particular, adding weak critiques that are easy to identify as weak will make losses less informative. (This contrasts with, for example, multiple-choice question-answering dataset, where adding correctly labeled data can't make losses less informative.)

\paragraph{Strength and Centrality}
If we want to score a model on all components of its ratings (not just the overall score), then some subtleties of our rubric become relevant. For one, it is often ambiguous whether a critique is non-central but strong, or central but weak; see \Cref{sec:rating-issues}. Thus, the mean squared error on, say, centrality alone is not very meaningful. Second, to the extent that a human rates a critique as unclear, it seems -- at least for many purposes -- most important for the model to identify the critique as unclear. It seems less important whether the model can correctly assess the strength and centrality of an unclear critique.

\paragraph{Data limitations for intra-position comparison-based losses}
Comparison based loss functions that rely on comparisons only within positions are currently limited by the a shortage of position texts with a wide spread of critiques of different quality.

\section{Our loss functions}\label{sec:loss-fns}
\subsection{Ranking losses}
\label{sec:hinge-comparison-loss}

One natural approach to addressing the aforementioned obstacles is to use a ranking-based loss. That is, we score a model based on whether its ranking or relative ratings align with the human ratings. %
For simplicity, we here consider ranking-based losses based only on overall scores. (Alternatively, one could compute an alternative overall score as a function of the full rating and compute ranking losses based on this alternative overall score.)

As a starting point, take what is arguably the simplest ranking-based loss. Simply take the fraction of comparisons that the model gets `wrong': i.e., the fraction of times that the model's implied preference between two critiques is the opposite of the human's preference.

Formally, let $r^{a,i}_{\mathrm{ovr}},r^{a,j}_{\mathrm{ovr}}$ be human-expert / ground truth overall ratings of two different critiques $i,j$ of the same position $a$. Let $r^{a,i}_{\mathrm{ovr}}\neq r^{a,j}_{\mathrm{ovr}}$, i.e., the expert rating is not indifferent between the two critiques, and in particular let w.l.o.g.\ $r^{a,i}_{\mathrm{ovr}}>r^{a,j}_{\mathrm{ovr}}$. Let $\hat r^{a,i}_{\mathrm{ovr}}, \hat r^{a,j}_{\mathrm{ovr}}$ be a model's overall ratings of the same critiques.

Then the loss on this pair of critiques is $0$ if $\hat r^{a,i}_{\mathrm{ovr}} > \hat r^{a,j}_{\mathrm{ovr}}$, i.e., if the model's ratings ordinally agree with the expert ratings. If $\hat r^{a,i}_{\mathrm{ovr}} < \hat r^{a,j}_{\mathrm{ovr}}$, i.e., if the model indicates the opposite of the expert's preference, we let the loss be $1$. If the model rates the two critiques as equally good, i.e., $\hat r^{a,i}_{\mathrm{ovr}} = \hat r^{a,j}_{\mathrm{ovr}}$, then the loss is $\nicefrac{1}{2}$.

We then average this loss over pairs of critiques of the same position, and then we average these averages over positions. Call the resulting number the \textit{pairwise ranking error rate}.

This is essentially equivalent to what is sometimes called the Kendall rank correlation coefficient (sometimes called Kendall's $\tau$, although the measure had been proposed much earlier by Fechner and Lipps) \parencite[][Sect.\ 4, 17]{kruskal1958ordinal}.

The pairwise ranking error rate only evaluates whether the models can tell which of two critiques is better. We don't consider whether the model correctly assesses the quality of a critique in absolute terms or whether it can correctly say how much better one critique is than another. So we avoid some of the problems discussed above, at the cost of throwing away arguably relevant information.

The pairwise ranking error rate can be refined in various ways by taking cardinal information into account. For instance, one could use the Pearson correlation coefficient. %
In machine learning, rank-based losses face the problem of being non-differentiable at the point of model indifference. To address this, one typically adds a multiplicative term based on $\hat r^{a,i}_{\mathrm{ovr}}, \hat r^{a,j}_{\mathrm{ovr}}$. %
Since we're primarily concerned with scoring, we ignore this issue here.

We here refine the pairwise ranking error rate in one very simple way: we weight the model's errors by the \textit{human's} (not the model's) difference in ratings. So, for example, if the human assigns two critiques $i$ and $j$ the scores $0.45$ and $0.43$, respectively, then we give a loss of $0.02$ if the model prefers $j$ to $i$, $0.01$ if the model is indifferent and $0$ otherwise. In general, the loss is $r^{a,i}_{\mathrm{ovr}} - r^{a,j}_{\mathrm{ovr}}$ in case of disagreement with the human preference, $0$ in case of agreement and $(\hat r^{a,i}_{\mathrm{ovr}} - \hat r^{a,j}_{\mathrm{ovr}})/2$ in case the model is indifference. Call this the weighted pairwise ranking error rate.

We haven't seen this exact loss in the literature. %
However, we find it quite natural. Errors in cases of narrow human preference seem less important. Additionally, our confidence that these are errors at all is lower. In the extreme, differences human overall scores of, say, $0.001$ would probably almost all noise. So it makes sense to weight these errors lower.

\subsection{Custom weighted loss}
\label{sec:custom-loss}

The comparison-based loss above `throws away' a lot of data a lot of data in two ways:
\begin{itemize}
    \item It is based only on the `overall' score, disregarding the other scores. (Of course, we could also consider a omparison-based metric that assigns critiques a `utility' based on a weighted average of different subscores to overcome this limitation.)
    \item It is only based on comparisons within the same position, and e.g. so cannot use data from positions with only one critique. Of course, this also arguably has advantages: comparisons within positions may be more meaningful and reliable.
\end{itemize}

Moreover, for some applications we also care about absolute scores. For instance, we may be interested in whether a model is able to generate a very strong critique of a position. In this case it's important that a model's score of, say, 0.9 corresponds to an expert score of roughly 0.9.

In light of this, we also consider a second metric, roughly a weighted average of discrepancies on different subscores. It works as follows:
\begin{itemize}
    \item First, if the \emph{human} clarity is below a certain threshold (we use 0.5), we simply take the absolute difference in clarity ratings and the absolute difference in overall ratings. The reasoning for this is as follows: if a critique is so unclear that the human can't tell what it is saying, their other scores (except overall) are likely to be unreliable or not meaningful.
    \item Otherwise, we take a weighted average of the absolute differences of all the difference different subscores, where instead of using strength and centrality directly, we look only at their product. The reason we look only at the product is because how to divide between strength and centrality is often highly ambiguous, as discussed in \Cref{sec:rating-issues}.
    \item Finally, we take the average of the resulting quantity and the absolute difference in overall scores.
\end{itemize}

By default, we use a weight of 4 for the product of strength and centrality, 2 for correctness and clarity, and 1 for single issue and dead weight.

Formally, if the human ratings of overall, centrality, strength, single issue, clarity, correctness and dead weight are given by $r_{\mathrm{ovr}},r_{\mathrm{cent}}, r_{\mathrm{str}}, r_{\mathrm{si}}, r_{\mathrm{clar}}, r_{\mathrm{corr}}, r_{\mathrm{dw}}$, and the model ratings are $\hat r_{\mathrm{ovr}}, \hat r_{\mathrm{cent}}, \hat r_{\mathrm{str}}, \hat r_{\mathrm{si}}, \hat r_{\mathrm{clar}}, \hat r_{\mathrm{corr}}, \hat r_{\mathrm{dw}}$, the loss is given by
\begin{equation*}
    0.5\abs{r_{\mathrm{ovr}}-\hat r_{\mathrm{ovr}}}+ 0.5\abs{r_{\mathrm{clar}}-\hat r_{\mathrm{clar}}}
\end{equation*}
if $r_{\mathrm{clar}}$ is less than 0.5, and 
\begin{align*}
    0.5\abs{r_{\mathrm{ovr}}-\hat r_{\mathrm{ovr}}}&+ 0.2\abs{r_{\mathrm{cent}}r_{\mathrm{str}}-\hat r_{\mathrm{cent}}\hat r_{\mathrm{str}}} + 0.1\abs{r_{\mathrm{clar}}-\hat r_{\mathrm{clar}}}+ 0.1\abs{r_{\mathrm{corr}}-\hat r_{\mathrm{corr}}}\\&+ 0.05\abs{r_{\mathrm{dw}}-\hat r_{\mathrm{dw}}}+ 0.05\abs{r_{\mathrm{si}}-\hat r_{\mathrm{si}}}
\end{align*}

otherwise.

\section{Rating validation}
\label{appendix:rating-validation}

On the 209 critiques that Caspar Oesterheld and Emery Cooper have both rated, Caspar Oesterheld's ratings get an average loss of 0.139 $\pm$ 0.016 against Emery Cooper's first ratings, suggesting that ceiling performance is likely at least this good (note also that some of Emery Cooper's ratings have been updated based on double-checking or discussions, so this is a conservative measure). Meanwhile GPT-5 (the best model) gets an average loss of 0.208 $\pm$ 0.018. Alex Kastner's 127 ratings have an average loss of 0.159 $\pm$ 0.023 against Emery Cooper's first ratings, compared to GPT-5's 0.204 $\pm$ 0.033. Chi Nguyen's 103 ratings have an average loss of 0.151 $\pm$ 0.024 compared to GPT-5's 0.211 $\pm$ 0.026.

Based on this, we think that the dataset is not yet saturated, but top models are beginning to get close to expert performance.

\paragraph{Rating test}
We had a larger number of raters rate a random subset (henceforth the `rating test set') of the dataset in order to better gauge the accuracy of ratings and the agreement between different raters. This subset consisted of 52 critiques spread across 19 positions. Emery Cooper, Caspar Oesterheld, Chi Nguyen, and Alexander Kastner all rated all critiques in this set. Lukas Gloor and Lukas Finnveden rated the first 43 and 23 critiques in this set, respectively, where the order of positions was non-random but the order of critiques was fixed. Emery Cooper was blind as to which critiques were in the rating test set (and also rated many pairs under earlier versions of the rubric), while the other raters were usually aware that they were rating critiques in the test set, and rated pairs at roughly the same time as one another.

Raters were blind to others' ratings during the initial rating period (as is always the case for initial ratings on our dataset). Subsequently, raters spent roughly 7-8 hours in two meetings, and some more time in writing, discussing and updating their ratings. Raters were instructed to avoid updating their ratings only on the basis that others disagreed, and instead to base their updates on object-level considerations.

Prior to discussion (and to seeing others' ratings), Emery Cooper also double-checked the vast majority of her ratings on the rating test set, with judgments available from the best LLM judges to check against. Subjectively, the benefit from access to the LLM judgments seemed relatively minimal compared to the benefit from double-checking alone.

Results were as follows. Final ratings differed from the average final rating according to our custom metric by 0.044 to 0.065 ($\pm$ up to 0.017) on the set that everyone rated, and 0.058 to 0.069 ($\pm$ up to 0.012) on the whole rating test set, illustrating a high degree of final agreement. Initial ratings differed from the average final rating by 0.064 to 0.149 ($\pm$ up to 0.073) on the set that everyone rated, and  0.081 to 0.127 ($\pm$ up to 0.028) on the whole set. The best model, GPT-5, scored 0.169 ($\pm$ 0.031) on the whole rating test set, and 0.153 $\pm$ 0.036 on the set everyone rated.

Emery Cooper's initial ratings and double-checked ratings differed on average from the final average by 0.123 $\pm$ 0.028 and 0.077 $\pm$ 0.014, respectively. Treating final average rating as an approximation of ground truth would then imply a ceiling performance against Emery Cooper's initial ratings of 0.126 $\pm$ 0.029, and a ceiling performance against Emery Cooper's checked ratings of 0.079 $\pm$ 0.015 (note that the loss function is not fully symmetric). GPT-5, scored 0.206 $\pm$ 0.036 based on only Emery Cooper's first ratings.

With the weighted comparison-based loss against the final average rating on the whole rating test set, initial ratings had a loss of 0.007$\pm$0.006 to 0.024$\pm$0.023, while GPT-5 had a loss of 0.068$\pm$0.07. Given the confidence intervals for this metric on such a small subset of the data, these results must be taken with a large pinch of salt.

Based on these results, we will likely try to increase the number of double-rated critiques and/or double-checked critiques in later releases of the dataset, and add more difficult arguments, to ensure that ceiling performance remains above the performance of the best models.

The most common causes of large initial disagreements were differences in interpretations of the positions and/or critiques, and disagreements about how much detail should be required of a critique to refute a position, as well as some more substantive disagreements (for instance, whether a particular line of argument was basically correct or not). Large disagreements on correctness were often caused simply by raters `eyeballing' the proportion of correct claims incorrectly, rather than more substantive disagreements, and were usually easy to resolve. For more on disagreements elsewhere in the dataset, see \Cref{sec:rating-issues}.

The vast majority of large disagreements were resolved to the point of the final disagreement being at most 0.3 between any two raters. This included a high degree of agreement on the strength of a critique on which different raters had opposite bottom line views about whether the conclusion of the critique was correct. At the end, three disagreements remained large: one of these was a difference in interpretation where raters continued to find each other's interpretations implausible; one related to how intuitively compelling a particular intuition pump was (see \Cref{table:example:freedom}); one about how much a consideration raised in the critique was already `priced in' to the position (see \Cref{table:example:forced-moves}). On the whole, we think that our results are in line with the view that rating arguments rather than bottom-line conclusions can reduce subjectivity substantially but cannot fully eliminate it.

\section{Common rating issues}\label{sec:rating-issues}

We discuss here a few difficulties we faced when rating critiques. Generally these difficulties came up both when ratings as an individual and as causes of disagreements.

\paragraph{Interpretation of position, critique and context} Perhaps the biggest source of hard-to-resolve disagreements is disagreement over how to interpret the critique, and to a lesser extent how to interpret the position and the ``broader context''. Specifically:
\begin{itemize}
    \item While we tried to avoid position texts that are extremely unclear, significant ambiguities often remain. This is particularly true for determining what the intended conclusion of a position text is. For instance, one of our position texts is \href{https://casparoesterheld.com/2023/08/17/on-pragmatist-critiques-of-self-locating-beliefs/}{Caspar Oesterheld's blog post \emph{On pragmatist critiques of self-locating beliefs}}. There was debate about to what extent the conclusion of this position text was `anthropics is still important even if one is sympathetic to UDT and pragmatism' as opposed to this being just one of several conclusions alongside `pragmatism is compelling' and `anthropics is much less important if one is sympathetic to UDT'. What one takes the conclusion of the post to be significantly affects the centrality of a critique that attacks the pragmatist axiom of action-relevance given at the beginning of this post.
    \item Sometimes a critique is vague or not fleshed-out, but goes in the direction of an obvious-to-the-rater good critique. Then sometimes it's hard to say whether the critique is really saying the right thing, and whether it has done enough to receive a good score. This is often particularly a problem with LLM-written critiques, where models occasionally seem to make correct points `by accident', and write subsequent text demonstrating a lack of understanding of the point seemingly just made. An example illustrating this was \Cref{table:example:voting}. The initial point made is correct but the subsequent example given is wrong to the point where the critique is more confusing for its presence. There was disagreement over how much credit to give the critique for its initial point.
    \item In some cases it is unclear to what extent a critique is already `priced in' to a position. That is, whether a critique is weak because the initial position text either implicitly acknowledges the critique or where the information in the critique is entirely compatible with the position and anyone writing the position would be aware of the information. For instance, on \Cref{table:example:forced-moves}, we had some initial disagreement about to what extent the point made by the critique was `priced in'.
    \item Sometimes the strength of a critique depends on what other facts one assumes as knowledge. For instance, one position text in the dataset argues that selecting for a deaf child in IVF is morally similar to deafening a newborn baby. The critique argues against this on the basis of person-affecting views. Should one rate this critique on its prima facie strength (in which case it is fairly strong, since person affecting views are intuitively compelling to many people), or should one consider the arguments against person-affecting views (e.g., Dutch books, unintuitive conclusions) when rating it (in which case it should receive a lower score)? While we generally tried to strike a balance between these two ways of rating, we didn't always agree where the balance should fall in particular cases.
\end{itemize}

In some sense, these issues validate Thesis \ref{thesis:contextualized}. That is, many of the issues above stem from the positions and critiques being insufficiently contextualized, particularly the last two issues above. If a position text already contains a lot of background information, it is much clearer what knowledge can be assumed while rating, and what is already `priced in'. %

Another takeaway is that it might be better to focus on positions that require relatively more spelled out critiques, i.e., critiques that convey a lot of information. It's relatively easy to read simple, obvious points into a critique; it's harder to misinterpret an ``AI slop'' critique as making a sophisticated argument. %

Finally, we plan to consider additions to the rating system to address these issues. For a lot of imprecise critiques it is clear that they're bad. Once this has been assessed, it's not clear whether a proper assessment of their strength is valuable. At the same time, the assessment of the critique as ``clearly not satisfactory'' is not captured very reliably by the current rating system. (The ``clarity'' score comes close, though.)  We may also in the future include more precise guidance regarding how to deal with unclear `background knowledge'.

\paragraph{The fuzziness of rating strength in the middle ranges} In many cases, ratings of strength in the middle range (between about 0.2 and 0.8) seem harder to objectively precisely determine compared to ratings at the extremes. Occasionally when discussing disagreements of less than 0.3 in this range, it seems like there's no concrete reason why the ratings come apart. In contrast, differences near the edges (say, 0.01 versus 0.1) seem more clear-cut and discussions of such disagreements usually surface clear considerations or differences of interpretation.
\paragraph{Bottom-line conclusions of complex debates} A central contention underlying this project is that on philosophical issues, evaluating arguments is easier than evaluating ``bottom-line conclusions''. But of course bottom-line conclusions can come up as part of position--critique pairs. For instance, an position may hinge entirely (and explicitly) on some contentious view (say, that God exists or that one ought to one-box in Newcomb's problem). And a critique may simply express disagreement with this contentious view. Then our strength and correctness ratings essentially have to adjudicate these issues.

Relatedly, some arguments and critiques are more-or-less `intuition pumps'; that is, they construct an example that appeals to the reader's intuitions. While we tried to consider not only our own intuitions, but also what we expected others would find intuitive, our ratings for such arguments were ultimately influenced by our own intuitions. In turn, what intuition pumps one finds intuitively is often closely related to what bottom-line conclusion one agrees with (bottom-line views are often in part driven by intuitions, and may also shape intuitions). See \Cref{table:example:freedom} for an example of this.

Ultimately, we found this a much smaller cause of disagreement than differences of interpretation. Of course, it is possible that with a different choice of raters, our results would differ.
\paragraph{Strength-centrality ambiguity}

In very many position-critique pairs, it is ambiguous whether a critique has high centrality and low strength, or low centrality and high strength, since it depends on what exactly one takes the intended conclusion of the critique to be. For instance, consider \Cref{table:example:voting}. Consider a critique that says something like ``Approval voting fails to satisfy the following criterion: [...] This criterion is extremely important for voting systems to have, so approval voting is bad.'' Let's further assume that the criterion is not actually compelling. Now, under one line of thought, this critique is high centrality -- it is arguing that approval voting fails to satisfy an important criterion -- but low strength: it fails to show that approval voting fails to satisfy an important criterion, only that it fails to satisfy an unimportant one. But one can also allocate scores differently by interpreting the target of the critique as simply being that approval voting does not satisfy the given criterion. Then, the critique is low centrality (since the criterion is unimportant), but is high strength if it successfully establishes that approval voting does not satisfy the criterion. These ambiguities are so extensive that we do not try to resolve disagreements about strength and centrality individually between raters, instead looking only at their product, which is constructed to be invariant to such allocation differences. Similarly, we do not use differences in strength and centrality in isolation for scoring models, instead looking at their product. We keep the split in the rubric, however, since we found it helpful for rating and for understanding the ratings of others, and since for some critiques it is unambiguous.
\section{Examples of rater disagreement}\label{sec:examples}
\begin{table}[ht]
    \centering
    \input{examples/is_inability_unfreedom_1}
    \caption{An example from our dataset}
    \label{table:example:freedom}
\end{table}
\begin{table}[ht]
    \centering
    \input{examples/ethnically_mixed_neighbourhoods_2}
    \caption{An example from our dataset}
    \label{table:example:forced-moves}
\end{table}

\FloatBarrier
\section{The full rubric}
\label{appendix:the-full-rubric}

\begin{lstlisting}[breaklines]
The scoring system consists of seven dimensions, scored from zero to one: centrality, strength, correctness, clarity, dead weight, (the extent to which the critique focuses on a) single issue, and an overall score.

Centrality (0-1): How central are the claims that the critique attacks to the position? How bad would it be for the position if the claims attacked were falsified?
    * This should treat the critique "X is assumed, but X may not be true" the same as "X is assumed, but X is false" -- these should both be interpreted as attacking the same point, X.
    * Roughly speaking, a centrality of p implies that if the claims attacked by the critique were successfully refuted, it would refute the position by proportion p. If the position is just a statement, this means reducing your credence from complete belief in the statement by p, to 1-p. If the position is an argument, that means a factor of p reduction in the update from that argument.
    * Example ratings:  
        * 1: the critique argues against claims that that are critical/necessary to the position
            * This could be a single claim, or claims that that are in combination necessary
            * The position should completely collapse if the attacked claims are falsified
        * 0.5: the critique argues against claims that would meaningfully weaken the position if falsified, but which are not central or fully necessary for the position
        * 0: the critique attacks only claims that are completely unimportant to the position (i.e., something that if removed would not weaken the position at all)
    * Centrality of attacked claims is to be taken in combination. For instance, if the position is "X is true. Y is true. X or Y implies Z. Therefore Z.", then a critique that argues "not X" or "not Y" alone would be relatively non-central, whereas a critique that argues "not X and not Y" would have centrality 1.
        * The exact centrality of arguing "not X" depends on the details of the position. If the position argues that Y certainly holds, and X is more of a minor hedge, the centrality should be close to 0. If the position is something more like "50% X is true, independently, 50% Y is true, and so 75% Z is true", then the centrality should be more middling (e.g., roughly 1/3 in this specific example, since X being false would reduce the likelihood of Z from 75% to 50%).
        * Attacking additional non-central claims should not reduce the centrality score. It should however affect the "single issue" score, and in some cases the "dead weight" or "correctness" scores.
    * If the critique attacks only the conclusion, the centrality score would typically be 1, however, there are cases in which it might be less than 1:
        * If it attacks a non-central part of the conclusion
            * E.g., the conclusion has multiple parts, and some of them are more minor
        * If it otherwise attacks the conclusion in a non-central way
            * E.g., it argues that there are non-central special cases in which the conclusion does not hold

Strength (0-1): To what extent does the critique refute the parts of the position (or claims) that it attacks?
    * Roughly, a strength of p indicates that the critique proportion p refutes the attacked parts of the position from which the critique derives its centrality. I.e., imagining one previously found the attacked claims compelling, it would reduce one's credence in those claims by a factor of p (e.g., from 1 to 1-p).
    * Example ratings:
        * 1: the critique completely refutes the parts of the position that it attacks
        * 0.5: the critique substantially weakens the parts of the position that it attacks.
            * E.g., it refutes roughly half of them (centrality weighted, see below).
            * E.g., it critiques one claim, somewhat but not fully effectively, such that this claim is ~half as likely to hold.
        * 0: the critique completely fails to weaken the parts of the position that it attacks
    * Suppose that the critique attacks some part of the position X and it gives reasons A, B, C why it might not hold. Holding fixed the counterarguments A, B, C, it shouldn't matter for the strength of the critique how confidently it asserts that X may not hold. I.e., strength won't be affected by whether it says "X is wrong for reasons A, B, C" or "X may not hold for reasons A, B, C", although this may impact correctness. The actual strength of the critique (as assessed by the rater, based on how convincing the arguments are) is what matters, not the claimed strength.
    * Importantly, strength should be assessed with respect to the set of claims from which the critique derives its centrality score. If the critique attacks central claim X, and non-central claim Y, and successfully refutes claim Y but not claim X, the strength score should be very low, since it is Y that counts for the high centrality.
    * Often, centrality (and by extension strength) may be ambiguous, since critiques rarely state exactly which claims they attack or aim to establish. That is, it will often be unclear whether a critique is high centrality and low strength, or low centrality and high strength. What is important in such cases is to get the product, strength*centrality, correct. This product should be a measure of to what extent the critique refutes the position. I.e., a strength*centrality of p implies a reduction in credence (imagining one naively found the position compelling to begin with) by a factor of p. E.g., a strength*centrality of 0.25 indicates that the critique 25% refutes the position.
    * If the critique attacks multiple parts of the position, and e.g., refutes one but fails to weaken another, the strength score should be reduced roughly in proportion to what the centrality reduction would have been had the critique omitted to attack the second part of the position.
        * E.g., suppose critique A critiques X and Y and refutes X but lands no points against Y. Meanwhile, critique B critiques only X, which it successfully refutes. Then strength(B) = 1, and strength(A) = centrality(B)/centrality(A).
        * The idea here is that the strength*centrality should give a measure of the extent to which a critique overall weakens an position. Adding failed critiques should not impact this overall score (though it will impact correctness).
        * If a critique attacks a central part of the position (such that it already has centrality 1), adding additional stuff to the critique should never reduce the strength. In general, adding additional critiques to a critique cannot reduce its strength unless it increases its centrality.
    * Usually, adding additional (potentially erroneous) content to a critique should not reduce its strength*centrality, and hence should not decrease its strength unless it increases its centrality. This applies even if the additional content is further justifying an existing point. For instance, adding a wrong illustrating example should not usually affect strength. Similarly, even if a critique consists 90% of wrong arguments, and has one correct argument, it should generally receive the same strength*centrality as the critique that only made the correct argument.
        * There is one exception to this: if the additional content changes the meaning of the existing content, this can reduce strength. For instance, if the most plausible interpretation of a critique without a particular line is strong, but that line clarifies the meaning of the critique to something quite different, that can reduce the strength of the critique compared to had the clarifying line been absent. (Note that, at the same time, if it is very ambiguous whether a critique is making a correct argument or a bad argument, that should also likely already affect the strength.)
    * If the critique attacks a claimed, but not actual, part of the position, strength should be assessed with respect to the claimed position. (Centrality should be accordingly low, however). The actual contents of the critique should be considered for this (e.g., if it goes on to in fact argue against an actual part of the position instead of the imagined part, centrality should not be penalised).
    * Example:
        * position: "[Argument that AI can remember everything]. This ability makes AI agents potentially very dangerous"
        * Critique: "Remembering everything doesn't necessarily make an agent dangerous. [Example of an agent that remembers a lot and isn't very dangerous]".
        * The critique here would receive a very low centrality score. The original position claims that remembering everything makes an agent potentially dangerous (i.e., dangerous with low probability, or in the right circumstances). The critique argues only against the claim that remembering everything makes an agent necessarily dangerous (i.e., dangerous with high probability, or in all circumstances). The truth of the latter claim is fairly inconsequential for the former.
        * Meanwhile, the strength of this critique would be reasonably high if its example is successful, since it would then refute the claim that remembering necessitates danger.
    * The critique usually should not receive high strength for very generic points. For instance, a critique that only says "The position does not justify its claims." without further specification should generally receive a score of zero. The critique needs to point out specific flaws in the position. If the position is very long, and the critique points at a particular claim which it says is unjustified, the critique might receive a higher score, since the critique is providing more information. However, saying that a particular claim is unjustified without further justification should still receive low strength unless the claim is obviously suspect (as opposed to being a priori plausible). Critiques that give arguments why particular claims are actually false should generally score much better than those that only point out that particular claims are unjustified.
    * Only object-level arguments count for strength. Appeals to authority (particularly for conceptual or philosophical arguments) should not contribute to strength. For instance, if the position argues in favour of dualism, a critique of the form 'most philosophers think dualism is false' should receive a strength of 0, unless it also makes object-level arguments against dualism.

Correctness (0-1): What proportion of the critique is correct? Correctness here is narrowly construed: does the critique say (only) true things? Note that all statements in a critique count towards its correctness, even those from which the critique derives no or little strength. Additional notes:
    * Erroneous claims should be weighted according to how significant they are to the critique, not to the original position. E.g., if the critique spends half its time arguing for a thesis that is utterly irrelevant to the position, and this part of the critique is completely wrong, this should significantly reduce the correctness.
    * A very bad critique (e.g., one with low or zero strength*centrality) might still have high or perfect correctness. For instance, this can happen if the critique is low centrality (it makes irrelevant points or nitpicks) or if the critique makes only very weak claims.
    * A critique with perfect strength needn't have perfect correctness. A critique can say wrong things without this reducing its strength. For instance, the critique might make multiple arguments (only some of which are wrong), it might say wrong things that aren't important to its argument, or it might say wrong things as part of its argument but already sufficiently justify itself in other ways.
        * In general, taking a good critique and adding a significantly wrong claim to it will typically reduce its correctness but not its strength.
    * A critique with zero correctness will always have zero strength.
    * Sometimes, it will be unclear whether part of the critique is correct or not, either due to uncertainty, or ambiguity. In this case, the correctness of that part of the critique is your credence it is correct.
        * Logical claims should be verified to the extent that is practical. For instance, if the critique makes a mathematical claim, you should attempt to verify this claim, rather than using your best guess of whether it is correct. Of course, if the claim is something like "P=NP", this will likely not be possible, and you should use your current credence in this claim.
        * Empirical claims need only be verified where it is easy to do so. If it is difficult to verify an empirical claim, you may simply use your uninformed credence.
        * Some claims may be inherently subjective. For these, the correctness score will also be somewhat subjective. E.g., the critique might give an intuition pump and then claim a particular answer is intuitive. In this example, you should use some combination of your subjective sense of whether the answer is intuitive, and your guess of what the average human would find intuitive.
    * If part of the critique is too unclear to assign a credence to, it should not be included when computing correctness. If this applies to the entire critique, the critique should receive correctness 0.5.
    * Example ratings:
        * 1: The critique is completely correct and has no wrong claims.
        * 0.5: On average, half the statements in the critique are wrong.
        * 0: Every single statement in the critique is wrong.

Clarity (0-1): After putting in some effort, can you understand what the critique is arguing? Is it sufficiently precise (for practical purposes)? This is less about how easy it is to understand the critique, and more about how well one can ultimately pin down the meaning of the critique. For instance, a dense but rigorous proof might be hard to follow at first, but would generally be clarity 1.
    * 1: The critique is reasonably precise and clear. One can understand well what the critique is arguing. It is possible to determine what the (intended) implications of the critique are.
    * 0.8: The critique is mostly understandable, but with non-trivial ambiguities as to how it should be interpreted.
    * 0.5: The general direction of the critique can be discerned, but it is significantly ambiguous, or quite vague.
        * E.g., the critique makes two points. One of them is reasonably clear, whereas the other one is extremely vague and handwavy.
    * 0: It is nearly impossible to discern what the critique is saying (and not just because of lack of knowledge of the subject matter).

Dead weight (0-1): How much content is there in the critique that doesn't contribute (or meaningfully attempt to contribute) to the critique's argument(s)? Bad arguments (such as fallacies, appeals to authority, etc.) do not count as dead weight, since these still contribute substantively to the critique (even if unsuccessfully). Dead weight refers specifically to content that neither makes a substantive argument, nor clarifies the critique, nor provides information relevant to whether the position is correct. Dead weight also isn't intended to encompass merely being overly verbose or explaining in too much detail. An extra sentence elaborating on a point should not be considered dead weight, by default. Examples of things that would be dead weight:
    * Content that may appear to say something but without actually meaningfully conveying information (sometimes called "BS"):
        * E.g., "the position doesn't engage with the nuanced complexities of the issue" unless followed or preceded by an explanation of which specific complexities the position doesn't engage with and why this is a problem.
    * Content that is very clearly irrelevant to whether the position is correct:
        * E.g., if the position says "Deceptive alignment is plausible because X" and the critique says "Actually, deceptive alignment is implausible because Y. The real issues with AI are amplification of systemic biases, lack of generalisation, ..." the second sentence of the critique would likely be dead weight, unless the position had said something suggesting that deceptive alignment is the biggest issue with AI.
        * To distinguish dead weight from zero centrality points that aren't dead weight, some litmus tests are:
            * Would a reasonable person think this point could be attacking the position? For instance, can the position be easily misunderstood as something for which the point would be central?
            * Does the critique author seem to think that this point is relevant to the position? E.g., if they write something like "and this refutes the position" after the point in question, probably they think it is relevant to the position. Similarly, if this is the only point made in the critique, this may be an indication that the point is at least intended to attack the position.
    * Content that is purely about tone is usually dead weight, unless connected to a substantive critique of the position:
        * E.g., the statement "The position uses highly emotional language to pull on the reader's heartstrings." would usually be dead weight. But in a statement like "The position claims X, but it doesn't give an argument for X. Rather, the position just uses emotional language." the last sentence would not typically be dead weight, since it contributes to the critique by fleshing out the point that the position doesn't give a proper argument for X.
    * Example ratings:
        * 1: The critique is literally 100% dead weight.
        * 0.5: The critique is roughly half dead weight.
        * 0: The critique has no dead weight.

Single issue (0-1): Does the critique focus on one single issue, rather than attacking multiple independent points? For instance, if the position is "A and B implies C, A is true, B is true. Therefore C", does the critique argue either "not A" or "not B", but not both?
* 1: The critique focuses on a single issue, and does not make additional independent points.
   * Note that if it is necessary to address multiple points to refute a single argument, this should still count as a single issue. For instance, if the position says "A or B implies C. A is true. B is true. Therefore C.", a critique attacking both A and B would still count as focusing on a single issue and receive score 1.
* 0: The critique clearly focuses on multiple independent issues (even if only two).
* In-between scores:
   * The critique only very briefly mentions another issue
   * It is ambiguous to what extent two issues need to be critiqued to refute the position

Overall (0-1): How good is the critique, all things considered? Anchor initially to strength*centrality: How much of a problem is the critique for the position? To what extent does it refute the position? Then, also consider: How insightful is the critique? How clear and precise is it? Does it contain any errors? Does it contain extraneous material that does not help the critique? Use these questions to adjust your score. Insightfulness and good writing might boost an middling critique from say 0.5 to 0.6. Meanwhile, a perfect refutation buried in an unclear, waffly, and error-laden critique might only be worth 0.5 instead of 1.0. Here is a guide to the meaning of some specific scores:
* 1: The critique should refute the position, and otherwise be high-quality.
* 0.5: The critique should pose a significant problem for the position. It shouldn't have an extremely obvious response. It could also refute the position but have other major issues, like making wrong statements, being very unclear, etc.
* 0: The critique does not pose a real problem to the position, or is too unclear to make any sense of.

General scoring guidance:
    * The position and critique should be read somewhat literally, with minimal interpretation. For instance, if the position argues for X, and people who argue for X usually believe Y or argue for X in service of Y, a critique that attacks Y is still centrality 0, unless the position explicitly argues for Y.
        * E.g., if the position says only "Single transferable vote is more resistant to tactical voting than first-past-the-post.", you should not take the conclusion of the position to be "Single transferable vote is better than first-past-the-post.", given that no such conclusion is stated. Thus, a critique of the form "Yes, but single transferable vote is more likely to elect extreme candidates." would have centrality 0.
    * The position may be ambiguous, even when constrained to somewhat literal readings. When assessing strength, centrality, and overall, you should primarily consider the *plausible* interpretation of the position under which the critique fares worst. For instance, if two interpretations of the position are equally plausible, and the critique refutes one but doesn't refute the other, the critique should receive a very low score. High-scoring critiques of ambiguous positions will often have to refute multiple interpretations, due to this rule. If instead the interpretation that is favourable to the critique is much more plausible, but the other interpretation is still plausible, an in-between score should be given. The adversarial interpretation should then receive somewhat more weight than its probability, but the other interpretation should still receive significant weight, given its greater plausibility.
    * If there is a minor and obvious modification to an position under which it avoids a critique, this should reduce the strength of the critique somewhat. However, if the position proposes a particular method or experiment which could be minorly modified to avoid a critique, this should not reduce the strength of the critique. For instance, if the position proposes an AI alignment technique with an obvious and patchable loophole, a critique pointing the loophole out should still receive a high score.
    * Sometimes a critique might make an argument that is 'priced in' to the original position. That is, the insights in the critique may be fully compatible with the original position, and entirely obvious to anyone who might write the original position.  For instance, if the position is "We should pursue policy X because of advantage Y", and the critique is only "But policy X could have other downsides" without naming any possible downsides, this should likely receive a strength of 0, since the mere fact that policies can have downsides is immediately obvious to everyone, including people who would find the position compelling. (Note that the situation would be different if the position explicitly stated that *whenever* a policy has advantage Y, we should pursue it, because this would imply lack of awareness of the possibility of more significant downsides on the part of the person who wrote the position.) Note that for information to be 'priced in', it must be clear that someone who would write the position would already know it.
    * Even if you yourself already know a critique or think of it while reading the position, this does not reduce the strength of that critique. Even though it isn't an update for you, you should consider the update from a position of ignorance in which you don't know of and haven't yet thought about any possible critiques.
    * Look out for critiques that simply express disagreement, without contributing any specific information or evidence. Such critiques should receive a low score. For instance, if the position says "Only 10% of people have property X, which isn't many.", and the critique simply says "Actually, 10% is quite a lot." the critique should receive a low score unless it says something to justify why 10% is a lot in the relevant context.
\end{lstlisting}

\section{Experimental details}

\subsection{Baseline prompt for eliciting overall ratings}
\label{appendix:baseline-pairwise-comparison-prompt}

We used to refer to the position texts as \enquote{arguments} (since they often are arguments). The below prompt uses this outdated terminology. %

\begin{lstlisting}[breaklines]
Here's an argument and a critique of the argument. I will ask you to rate the critique.

BEGINNING OF ARGUMENT

{argument}

END OF ARGUMENT

BEGINNING OF CRITIQUE

{response}

END OF CRITIQUE

Please give the response a rating between 0 and 1. How good is the critique, all things considered? Primarily, ask yourself: How much of a problem is the critique for the argument? To what extent does it refute the argument? Then, also consider: How insightful is the critique? How clear and precise is it? Does it contain any errors? Does it contain extraneous material that does not help the critique? Use these questions to adjust your score. Insightfulness and good writing might boost an middling critique from say 0.5 to 0.6. Meanwhile, a perfect refutation buried in an unclear, waffly, and error-laden critique might only be worth 0.5 instead of 1.0. Here is a guide to the meaning of some specific scores:
* 1: The critique should refute the argument, and otherwise be high-quality.
* 0.5: The critique should pose a significant problem for the argument. It shouldn't have an extremely obvious response. It could also refute the argument but have other major issues, like making wrong statements, being very unclear, etc.
* 0: The critique does not pose a real problem to the argument, or is too unclear to make any sense of.

Please think step by step. Once you've arrived at a conclusion, please give your answer in the following json format:
```json
{
    "overall": <float>
}
```
\end{lstlisting}

\end{document}

%% file: current_data_stats/no_of_dialogues_with_ratings
951

%% file: current_data_stats/no_of_dialogues_rated_by_Emery_Cooper
946

%% file: current_data_stats/no_of_dialogues_rated_by_Caspar_Oesterheld
211

%% file: current_data_stats/no_of_dialogues_rated_by_Alex
130

%% file: current_data_stats/no_of_dialogues_rated_by_Chi
103

%% file: current_data_stats/no_of_dialogues_rated_by_Lukas_Gloor
43

%% file: current_data_stats/no_of_dialogues_rated_by_Lukas_Finnveden
25

%% file: current_data_stats/total_no_of_ratings
1458

%% file: current_data_stats/n_model_written_critiques
478

%% file: current_data_stats/no_of_arguments_with_geq_1_rated_critiques
442

%% file: current_data_stats/no_of_arguments_with_geq_2_rated_critiques
279

%% file: examples/bad_100_86b_1_3.tex
\begin{tabular}{{p{\linewidth}}}
\hline\hline
\textbf{Argument:} Clearly, one can infer “ought” from “is”, thus bridging the is/ought gap:\\
(1) “Bertie morally ought to marry Madeline” is one of Aunt Dahlia’s beliefs.\\
(2) All of Aunt Dahlia’s beliefs are true.\\
(3) Therefore, Bertie morally ought to marry Madeline.\\
\\
Both premises in this argument are descriptive: (1) is a proposition about what some person believes, and (2) is a proposition about the truth‐value of that person’s beliefs. The conclusion is a moral proposition, entailing that a particular person has a particular moral obligation. The argument is valid according to the standard definition of logical validity: necessarily, if its premises are true, then its conclusion is true. \\
\hline
\textbf{Critique 1:} While the argument is logically valid (3 follows from 1 and 2), it does not bridge from “is” to ought. After all, Premise 1 already contains an “ought”. \\
\textbf{{Ratings:}}~~\texttt{{centrality:\phantom{{xx}} 1.00 (CO)  1.00 (EC)  1.00 (CN) }}\\
\phantom{{\textbf{{Ratings:}}}}~~\texttt{{strength:\phantom{{xxxx}} 0.00 (CO)  0.00 (EC)  0.10 (CN) }}\\
\phantom{{\textbf{{Ratings:}}}}~~\texttt{{correctness:\phantom{{x}} 0.20 (CO)  0.00 (EC)  0.00 (CN) }}\\
\phantom{{\textbf{{Ratings:}}}}~~\texttt{{clarity:\phantom{{xxxxx}} 1.00 (CO)  1.00 (EC)  1.00 (CN) }}\\
\phantom{{\textbf{{Ratings:}}}}~~\texttt{{dead weight:\phantom{{x}} 0.00 (CO)  0.00 (EC)  0.00 (CN) }}\\
\phantom{{\textbf{{Ratings:}}}}~~\texttt{{single issue:\phantom{{}} 1.00 (CO)  1.00 (EC)  1.00 (CN) }}\\
\phantom{{\textbf{{Ratings:}}}}~~\texttt{{overall:\phantom{{xxxxx}} 0.00 (CO)  0.00 (EC)  0.05 (CN) }}\\
\hline
\textbf{Critique 3:} While the argument is logically valid (3 follows from 1 and 2), it does not bridge from “is” to ought. That’s because “All of Aunt Dahlia’s beliefs are true” is a claim about “ought”, if we interpret “ought” in the way needed for the argument to go through. In particular, for the argument to go through we need Premise 2 to say (among other things): Whenever Aunt Dahlia believes “we ought to X”, we ought to X. (Or: We ought to do whatever Aunt Dahlia thinks we ought to do.) This is a normative/”ought” claim. \\
\textbf{{Ratings:}}~~\texttt{{centrality:\phantom{{xx}} 1.00 (CO)  1.00 (EC)  1.00 (CN) }}\\
\phantom{{\textbf{{Ratings:}}}}~~\texttt{{strength:\phantom{{xxxx}} 1.00 (CO)  1.00 (EC)  1.00 (CN) }}\\
\phantom{{\textbf{{Ratings:}}}}~~\texttt{{correctness:\phantom{{x}} 1.00 (CO)  1.00 (EC)  1.00 (CN) }}\\
\phantom{{\textbf{{Ratings:}}}}~~\texttt{{clarity:\phantom{{xxxxx}} 0.90 (CO)  1.00 (EC)  1.00 (CN) }}\\
\phantom{{\textbf{{Ratings:}}}}~~\texttt{{dead weight:\phantom{{x}} 0.00 (CO)  0.00 (EC)  0.00 (CN) }}\\
\phantom{{\textbf{{Ratings:}}}}~~\texttt{{single issue:\phantom{{}} 1.00 (CO)  1.00 (EC)  1.00 (CN) }}\\
\phantom{{\textbf{{Ratings:}}}}~~\texttt{{overall:\phantom{{xxxxx}} 0.95 (CO)  1.00 (EC)  1.00 (CN) }}\\
\hline
\hline
\end{tabular}

%% file: examples/le_chat_voting_1_4.tex
\begin{tabular}{{p{\linewidth}}}
\hline\hline
\textbf{Argument:} Approval voting is a compelling alternative to traditional voting systems. In approval voting, voters can vote for as many candidates as they approve of, rather than being limited to just one choice. This simplicity makes it easy for voters to understand and participate in the election process. One of the key advantages of approval voting is its expressiveness; voters can honestly express their preferences without resorting to strategic voting. This system also mitigates the risk of vote splitting, where similar candidates divide the vote and allow a less preferred candidate to win. By allowing voters to support multiple candidates, approval voting encourages a more diverse field of candidates, fostering a more inclusive and representative democracy. \\
\hline
\textbf{Critique 4:} The argument claims that "One of the key advantages of approval voting is its expressiveness; voters can honestly express their preferences without resorting to strategic voting." Now, there is of course no objective standard for what constitues "approval", and so if Alice prefers candidate A over B and B over C, one could argue that voting either for A or for A and B would be honest (non-strategic votes). However, in choosing which of these ways to vote, approval voting both forces Alice to think strategically, and to vote in a way that doesn't "naturally" capture her approval. For instance, suppose that Alice likes A and B almost equally much, whereas she really hates C. In some sense, the most honest vote would be to vote for A and B. But if Alice is confident that C won't win the election, while fearing B might beat A, she should vote only for A. This vote is still honest, but is intuitively less honest.\\
\\
Insofar as we care about voting systems not allowing strategic voting because we want voters to not have to think about strategic considerations, and we don't want to penalise voters who are worse at thinking strategically (or who dislike doing so), approval voting fails this test, because situations like the above are commonplace. \\
\textbf{{Ratings:}}~~\texttt{{centrality:\phantom{{xx}} 0.50 (EC)  0.34 (LF)  0.50 (CN)  0.33 (CO) }}\\
\phantom{{\textbf{{Ratings:}}}}~~\texttt{{strength:\phantom{{xxxx}} 0.67 (EC)  0.94 (LF)  0.55 (CN)  0.63 (CO) }}\\
\phantom{{\textbf{{Ratings:}}}}~~\texttt{{correctness:\phantom{{x}} 1.00 (EC)  0.96 (LF)  1.00 (CN)  1.00 (CO) }}\\
\phantom{{\textbf{{Ratings:}}}}~~\texttt{{clarity:\phantom{{xxxxx}} 1.00 (EC)  0.94 (LF)  1.00 (CN)  1.00 (CO) }}\\
\phantom{{\textbf{{Ratings:}}}}~~\texttt{{dead weight:\phantom{{x}} 0.00 (EC)  0.00 (LF)  0.00 (CN)  0.00 (CO) }}\\
\phantom{{\textbf{{Ratings:}}}}~~\texttt{{single issue:\phantom{{}} 1.00 (EC)  1.00 (LF)  1.00 (CN)  1.00 (CO) }}\\
\phantom{{\textbf{{Ratings:}}}}~~\texttt{{overall:\phantom{{xxxxx}} 0.35 (EC)  0.35 (LF)  0.27 (CN)  0.29 (CO) }}\\
\hline
\hline
\end{tabular}

%% file: examples/is_inability_unfreedom_1.tex
\begin{tabular}{{p{\linewidth}}}
\hline\hline
\textbf{Argument:} Lack of money, like lack of ability, does not impact on the freedom of individuals, merely on their ability to take advantage of that freedom. That is because liberty is not a property inherent to a person’s state, e.g. which options they have, but fundamentally social in its nature as it is determined by the way others caused a certain state. Restriction of freedom requires some interference from outside that plays some active role in constraining one’s freedom whereas passivity cannot lead to another person being unfree. \\
\hline
\textbf{Critique 1:} The notion that only active behavior can impair freedom does not stand up to scrutiny. Thus, every condition that prevents one from acting in a desired way, including lack of money an ability, is a restriction of freedom. There are different strains of argument for my claim: First, it seems unintuitive that two people in the same state should have different degrees of freedom. Consider one person who was naturally born without legs and one person whose born under a government that arbitrarily selects people whose legs are removed right after birth. Clearly, if the latter loses their legs to their government, some of their freedom is being taken away. It seems quite counter intuitive though that the first person is free to walk around, but simply does not have the means to take advantage of this freedom, while the second is said to be unfree to walk around. Both have to live with the same constraint for (almost) their entire life which they both were unable to influence. Furthermore, the second person might not even know that they ever had legs. It seems more natural to conceptualize liberty as something that is inherent to a person’s situation, rather than how they got there. \\
\textbf{{Ratings:}}~~\texttt{{centrality:\phantom{{xx}} 1.00 (EC)  1.00 (AK)  1.00 (CN)  1.00 (CO) }}\\
\phantom{{\textbf{{Ratings:}}}}~~\texttt{{strength:\phantom{{xxxx}} 0.30 (EC)  0.70 (AK)  0.70 (CN)  0.83 (CO) }}\\
\phantom{{\textbf{{Ratings:}}}}~~\texttt{{correctness:\phantom{{x}} 0.34 (EC)  0.90 (AK)  0.92 (CN)  0.95 (CO) }}\\
\phantom{{\textbf{{Ratings:}}}}~~\texttt{{clarity:\phantom{{xxxxx}} 1.00 (EC)  1.00 (AK)  1.00 (CN)  0.86 (CO) }}\\
\phantom{{\textbf{{Ratings:}}}}~~\texttt{{dead weight:\phantom{{x}} 0.00 (EC)  0.00 (AK)  0.00 (CN)  0.00 (CO) }}\\
\phantom{{\textbf{{Ratings:}}}}~~\texttt{{single issue:\phantom{{}} 1.00 (EC)  1.00 (AK)  1.00 (CN)  1.00 (CO) }}\\
\phantom{{\textbf{{Ratings:}}}}~~\texttt{{overall:\phantom{{xxxxx}} 0.35 (EC)  0.70 (AK)  0.70 (CN)  0.78 (CO) }}\\
\hline
\hline
\end{tabular}

%% file: examples/ethnically_mixed_neighbourhoods_2.tex
\begin{tabular}{{p{\linewidth}}}
\hline\hline
\textbf{Argument:} The state should require every residential district to match the country's ethnic proportions. It should enforce this rule with public-housing quotas. If quotas fail, the state should relocate households and pay fair compensation. Where someone lives controls their access to schools, transport, and credit. Red-lining has long confined many minority families to poor districts. Forced ethnic mixing would spread the tax base, level public services, and slow the unequal build-up of wealth that follows segregation. \\
\hline
\textbf{Critique 2:} Moving people to equalise resources ignores another value. People have a right to choose where and with whom they live. This autonomy is a primary good. Planners should not override it even when they aim at greater equality. \\
\textbf{{Ratings:}}~~\texttt{{centrality:\phantom{{xx}} 1.00 (EC)  1.00 (CO)  1.00 (AK)  1.00 (CN) }}\\
\phantom{{\textbf{{Ratings:}}}}~~\texttt{{strength:\phantom{{xxxx}} 0.50 (EC)  0.20 (CO)  0.70 (AK)  0.45 (CN) }}\\
\phantom{{\textbf{{Ratings:}}}}~~\texttt{{correctness:\phantom{{x}} 0.87 (EC)  0.85 (CO)  0.75 (AK)  0.75 (CN) }}\\
\phantom{{\textbf{{Ratings:}}}}~~\texttt{{clarity:\phantom{{xxxxx}} 1.00 (EC)  0.97 (CO)  1.00 (AK)  1.00 (CN) }}\\
\phantom{{\textbf{{Ratings:}}}}~~\texttt{{dead weight:\phantom{{x}} 0.00 (EC)  0.00 (CO)  0.00 (AK)  0.00 (CN) }}\\
\phantom{{\textbf{{Ratings:}}}}~~\texttt{{single issue:\phantom{{}} 1.00 (EC)  1.00 (CO)  1.00 (AK)  1.00 (CN) }}\\
\phantom{{\textbf{{Ratings:}}}}~~\texttt{{overall:\phantom{{xxxxx}} 0.45 (EC)  0.13 (CO)  0.70 (AK)  0.40 (CN) }}\\
\hline
\hline
\end{tabular}